\documentclass[12pt,a4]{article}

\usepackage{amsmath}
\usepackage{epsfig,amsthm}
\usepackage{amssymb,amsfonts}
\usepackage{dsfont}
\usepackage{graphicx}
\usepackage{subfigure}
\usepackage{indentfirst}
\usepackage{mathrsfs}
\usepackage{url}
\usepackage{longtable}
\usepackage{xparse}
\usepackage{bm}
\usepackage{bbm}
\usepackage{siunitx}
\usepackage{cite}
\usepackage{authblk}
\usepackage{color}
\usepackage{cleveref}
\usepackage{booktabs}
\usepackage{tabularx}
\usepackage{amsthm,amssymb}

\NewDocumentCommand{\tensor}{t_}
 {%
  \IfBooleanTF{#1}
    {\tensop}
    {\otimes}%
 }
\NewDocumentCommand{\tensop}{m}
 {%
  \mathbin{\mathop{\otimes}\displaylimits_{#1}}%
 }

\providecommand{\keywords}[1]{\textbf{\textit{Keywords---}} #1}
\newcommand{\abs}[1]{\lvert#1\rvert}
\newtheorem{theorem}{Theorem}

\title{Energy stable neural networks for gradient flow equations}
\begin{document}

\author[a]{Yue Wu}
\author[a]{Tianyu Jin}
\author[a]{Chuqi Chen\thanks{Corresponding author. E-mail address: cchenck@connect.ust.hk (C. Chen).}}
\author[a]{Ganghua Fan\thanks{Corresponding author. E-mail address: gfanab@connect.ust.hk (G. Fan).}}
\author[a]{Yuan Lan}
\author[c]{Luchan Zhang \thanks{Corresponding author. E-mail address: zhanglc@szu.edu.cn (L.Zhang).}}
\author[a,b]{Yang Xiang \thanks{Corresponding author. E-mail address: maxiang@ust.hk (Y.Xiang).}}

\affil[a]{Department of Mathematics, The Hong Kong University of Science and Technology, Clear Water Bay, Hong Kong Special Administrative Region of China}

\affil[b]{Algorithms of Machine Learning and Autonomous Driving Research Lab, HKUST Shenzhen-Hong Kong Collaborative Innovation Research Institute, Futian, Shenzhen, China}

\affil[c]{School of Mathematical Sciences, Shenzhen University, Shenzhen 518060, China}

\renewcommand*{\Affilfont}{\small\it}
 \date{}
\maketitle

\begin{abstract}

  We propose an 
    energy stable network (EStable-Net) for solving gradient flow equations.
      The EStable-Net enables decreasing of a discrete energy along the neural network, which is consistent with the property  of the gradient flow equation.
     The architecture of the neural network EStable-Net is based on the block network structure (Autoflow) in which output of each block can be interpreted as an intermediate state of the evolution process of the equation, and the energy stable property
     is incorporated in each block, which is easily generalized to include other physical and/or numerical properties. Our EStable-Net is a supervised learning network approach for solving evolution equations which does not depend on the convergence of time step goes to $0$, and can be applied generally even when only data is available but the equation is unknown. 
     We also propose a training strategy for supervised learning  that employs data of the evolution stages with different nature.
The EStable-Net is validated by numerical experimental results based on the Allen-Cahn equation and the Cahn-Hilliard equation in two dimensions.




\end{abstract}


\section{Introduction}

Partial differential equations are important tools in solving a wide range of problems in science and engineering fields.
 Over the past twenty years, deep neural networks (DNNs) \cite{Goodfellow2016,LeCun2015} have demonstrated their power in science and engineering applications, and efforts have been made to employ DNNs to solve complex partial differential equations as an alternative to the traditional numerical schemes.

%



Early works  \cite{lagaris1998artificial, dissanayake1994neural} use feedforward neural network to learn the initial/boundary value problem by constraining neural networks using differential equation.
Methods using continuous dynamical systems to model high-dimensional nonlinear functions used in machine learning were proposed in
\cite{e2017proposal}. A deep learning-based approach to solve high dimensional parabolic partial differential equations (PDEs) based on the formulation of stochastic differential equations was developed in \cite{han2018solving}.
Benefiting from autograd differentiation, physics-informed neural network (PINN) \cite{raissi2019physics} was derived for a wide variety of PDEs, using information of the PDE with initial/boundary conditions and a few time-space points randomly sampled in the solution domain to approximate the solution and trained by minimizing the mean squared error loss of the PDE.
The Deep Ritz Method \cite{yu2018deep} was proposed based
on representing the trial functions by deep
neural networks, which is in the variational form with the total energy as the loss.
%
The Deep Galerkin method (DGM) was proposed in \cite{sirignano2018dgm} with the PDE as the loss for the neural network approximation of the solution. The Weak Adversarial Network (WAN) \cite{zang2020weak} solves a PDE by a generative adversarial network using the weak form of the PDE as the loss. In \cite{lyu2022mim}, a deep mixed residual method (MIM) is proposed to solve high order PDEs by rewriting the original equation into a first-order system and taking the first-order system's residual as the loss function. The random feature method (RFM) \cite{chen2022bridging,chen2023rfm} and randomized neural network method \cite{rnn1} were proposed  and analyzed \cite{chen2024quantifying} for solving PDEs 
The Friedrichs learning \cite{chen2023friedrichs} employs Friedrichs minimax formulation as the loss to approximate solution of PDEs. In these methods, neural networks are trained for individual PDE problems, i.e., for different PDE problems, different neural networks need to be trained.


Another class of approaches  employ neural networks to learn the operators. The Fourier Neural Operator (FNO) was developed in \cite{li2020fourier}, which focuses on mappings between infinite-dimensional function spaces by using Fourier Transform to approximate the linear or nonlinear operators.
The Deep Operator Network (DeepONet) proposed in \cite{lu2021learning} learns nonlinear operators associated with
PDEs from data based on the approximation theorem for operators by neural networks.
In \cite{khoo2021solving},  parametric PDE problems with artificial neural networks were solved by expressing the physical quantity interested as a function of random coefficient.
A compact network architecture to solve electrical impedance tomography problem
based on the numerically low-rank property of the forward and inverse maps was proposed in \cite{fan2020solving}.
There are methods based on learning Green’s functions for solving PDEs \cite{deepgreen2020,zhang2021mod,hao2024multiscale}.
A neural network to approximate functionals without curse of dimensionality based on the method of Barron space was developed in
\cite{yang2022dimension}. A feature flow regularization method for improving structured sparsity in deep neural network was proposed in \cite{ffr}. Multigrid based networks for solving
parameterized partial differential equations were proposed in \cite{chen2022meta,HeJuncai2023}.
A multiscale operator learning method with hierarchical attention was developed \cite{ZhangLei2024}.
In \cite{liu2022deeppropnet}, the DeepPropNet was developed to approximate the evolution operator over a long time by using a single neural network propagator recursively.
Based on the operator splitting method, \cite{lan2023dosnet} proposed the Deep Operator-Splitting Network (DOSnet) for solving evolution equations. This network takes the form of block structure and
each block models the transition between the states in the physical evolution systems rather than mapping the input into some higher dimensional features, and the properties of the PDE is reflected in the network by using
the nonlinear part of the operator as the activation function.
Another category of operator learning methods focus on learning the numerical discretization of operators and leverages the computational advantages of neural networks to accelerate the process \cite{ZHOU2023112341,JuLili2024,jin2024fast,chen2024automatic}.
Many of these proposed neural network methods for solving PDE operators are based on the supervised learning trained from the available data \cite{li2020fourier,lu2021learning,khoo2021solving,fan2020solving,liu2022deeppropnet,lan2023dosnet,WangQi2023}.
One important research topic is to encode physical properties into the neural network \cite{CL2024dynamic,goswami2023physics, long2018pde}.


A class of widely used mathematical models take the form of gradient flows to minimize some total energy, for which many  numerical methods have been developed to preserve the energy dissipation property \cite{Elliott1993,Eyre1998,Shen1999,wise2009energy,shen2010numerical,wang2010unconditionally,chen2012linear,shen2012second,EnergyStable2013,yang2016linear,yang2017numerical,yang2017numerical2,shen2018scalar,pan2023novel}.
In this paper, we propose an energy stable neural network (EStable-Net) to solve the gradient flow equations. Based on the block architecture Autoflow of the DOSNet \cite{lan2023dosnet},
we introduce  auxiliary variables and encoded it into the  blocks
to enforce the  energy decay property along the network, which is consistent with the physical property of the gradient flow. This design provides a stable, efficient and interpretable network structure. The network is trained at a finite time $T$ for varying initial date, and once it trained, it can be applied to predict the evolution result at the training time $T$ and to infer the evolution results at later time.
Experiments on based on the Allen-Cahn equation and the Cahn-Hilliard equation in two dimensions demonstrate that our network is able to generate high accuracy and stable predictions.

Recently, some neural network based methods have been proposed to incorporate the physics of the solutions of the PDEs based on numerical schemes.
The unsupervised deep learning method in \cite{JuLili2024} trained a neural network that uses the loss of residual of the fully-discrete system  (Crank-Nicolson scheme of the Allen-Cahn equation) instead of using residual of the equation as the loss in the PINN method \cite{raissi2019physics}, which requires time step $\Delta t\rightarrow 0$; and their network is applied to both the classic and
conservative Allen-Cahn equations. The energy-dissipative evolutionary deep operator neural network in \cite{ShenJie2024} trained the DeepONet  \cite{lu2021learning} for the evolution over a time period $\Delta t$, and then in the subsequent evolution with time step $\Delta t$,
the network parameters are updated by a finite difference scheme based on the SAV  (scalar auxiliary variable) form the equation \cite{shen2018scalar} that enables the energy-dissipative property; this network also requires time step $\Delta t\rightarrow 0$.

Our EStable-Net is a supervised learning network approach for a gradient flow equation, which can be applied generally even when only date is available but the equation is unknown. The network structure of the EStable-Net that enables the energy stable property does not depend on the specific form of the PDE, unlike most of the available neural networks for solving PDEs (e.g., \cite{yu2018deep,raissi2019physics,sirignano2018dgm,JuLili2024,ShenJie2024}). In our EStable-Net, the time increment is finite and the time step $\Delta t\rightarrow 0$ is not required, unlike in \cite{JuLili2024,ShenJie2024} discussed above and many other available network methods for solving evolution equations; this leads to an efficient network and a fast inference of the evolution result after finite time once the network is trained, which fully takes the advantage of the neural network methods over the traditional numerical methods. The energy stable property of the EStable-Net suggests that the predicted evolution is not far from the actual evolution path, which reflects the physics of the PDE. Finally, the block network structure Autoflow can be applied generally to incorporate other physical and/or numerical properties beyond the energy stable property.

We also propose the following training strategy for supervised learning based on the physics of the gradient flow dynamics that incorporates the evolution stages with different nature. For example,  the dynamics of the Allen-Cahn equation and the Cahn-Hilliard equation consist of two evolution stages of phase separation and coarsening
We use both types of phase separation and coarsening date in the training, which has been shown to be able to significantly increase the accuracy of the network predictions in the training and the inference for later evolution.

The rest of this paper is organized as follows. In Section~\ref{sec:gradient-flow}, we
 briefly review the gradient flow equation and its energy dissipation property,
 which will inspire our energy stable neural network EStable-Net. In Section~\ref{energy_decay_formula},
  we present our energy stable neural network EStable-Net, which is
based on the block architecture Autoflow of the DOSNet \cite{lan2023dosnet}
and enforces the  energy stable property along the network.
In Section~\ref{sec:numerical}, we present some numerical experiments based on the Allen-Cahn equation and the Cahn-Hilliard equation in two dimensions to validate our EStable-Net.  We also propose a training strategy for supervised learning  that employs data of the evolution stages with different nature.
In Section~\ref{sec:dis}, we further examine the energy stable network design and the physics based two-stage training strategy.
We conclude this paper in Section~\ref{sec:conclusion}. An alternative formulation that enforces decrease of a discrete version of the energy in an energy stable block is presented in Appendix~\ref{app:A}.

\section{Gradient flow equations and energy dissipation property}\label{sec:gradient-flow}

In this section, we first briefly review the gradient flow equation and its energy dissipation property, which  suggests the neural network to be presented in the next section.


We consider the general form of gradient flow equations:
\begin{align}
    &\phi_t = -\mathcal{G}(\mathcal{D}\phi+f(\phi)), \quad (x,t)\in \Omega\times[0,T]\label{eq..gradflow}\\
    &\phi(0,x) = \phi_0(x),\label{eq..gradflowi} \quad x\in \Omega.
\end{align}
where 
$\mathcal{G}$ and $\mathcal{D}$ are positive operators. For the boundary conditions, we start with the periodic boundary conditions  for simplicity in the presentation of our neural network structure.

The energy of the gradient flow equation \eqref{eq..gradflow} is
\begin{equation}\label{eq..contenergy}
    E(\phi) = \int_\Omega \left(\frac12 \abs{\mathcal{D}^{1/2}\phi}^2+F(\phi)\right)\mathrm{d}x,
\end{equation}
where $ F'(\phi) =f(\phi)$ in \eqref{eq..gradflow}. With this energy functional, the gradient flow in Eq.~\eqref{eq..gradflow} can be written as
\begin{equation}
    \phi_t = -\mathcal{G}\left( \frac{\delta E}{\delta \phi}\right),
\end{equation}
where
\begin{equation}
\frac{\delta E}{\delta \phi}=\mathcal{D}\phi+f(\phi)
\end{equation}
 is the functional derivative of the energy $E$. Under this gradient flow equation, the energy \eqref{eq..contenergy} dissipates during the evolution:
    \begin{flalign}\label{eqn:energy-diss}
    \frac{\mathrm{d}}{\mathrm{d} t} E(\phi(t)) = \left(\frac{\delta E}{\delta\phi}, \phi_t\right) 
     = - \left\|\mathcal{G}^{1/2}\left(\frac{\delta E}{\delta \phi}\right)\right\|^2= - \|\mathcal{G}^{-1/2}\left(\phi_t\right)\|^2\le 0.
\end{flalign}
Here $\mathcal{D}^{1/2} (\mathcal{D}^{1/2})^* = (\mathcal{D}^{1/2})^* \mathcal{D}^{1/2} = \mathcal{D}$, and $\mathcal{G}^{1/2} (\mathcal{G}^{1/2})^* = (\mathcal{G}^{1/2})^* \mathcal{G}^{1/2} = \mathcal{G}$.

    Some examples of the gradient flow equations include:

         (i) The Allen-Cahn equation \cite{AC}, $\mathcal{G} = 1, \mathcal{D} = -\epsilon^2 \Delta, f(\phi) = \phi^3-\phi$, that is
        \begin{equation}\label{eqn:AC}
      {\rm (AC):} \ \ \   \ \ \ \ \  \phi_t = \epsilon^2 \Delta\phi-\phi^3+\phi.
        \end{equation}
        Its energy dissipation rate is
      \begin{flalign}\label{eqn:energy-diss-ac}
    \frac{\mathrm{d}}{\mathrm{d} t} E(\phi(t)) = \left(\frac{\delta E}{\delta\phi}, \phi_t\right)
     = - \left\|\frac{\delta E}{\delta \phi}\right\|^2= - \|\phi_t\|^2.
\end{flalign}

        (ii)  The Cahn-Hilliard equation \cite{CH1,CH2}, $\mathcal{G} = -\Delta, \mathcal{D} = -\epsilon^2 \Delta, f(\phi) = \phi^3-\phi$, that is
       \begin{equation}\label{eqn:CH}
    {\rm (CH):} \ \ \         \phi_t = \Delta(-\epsilon^2 \Delta\phi+\phi^3-\phi).
        \end{equation}
    Its energy dissipation rate is
   \begin{flalign}\label{eqn:energy-diss-ch}
    \frac{\mathrm{d}}{\mathrm{d} t} E(\phi(t)) = \left(\frac{\delta E}{\delta\phi}, \phi_t\right)
     = - \left\|\nabla\frac{\delta E}{\delta \phi}\right\|^2.
\end{flalign}

     Here, for the Allen-Cahn equation and the Cahn-Hilliard equation, the energy is
     \begin{equation}\label{eqn:e-ac-ch}
     E(\phi) = \int_\Omega \left(\frac{\varepsilon^2}{2}\abs{\nabla\phi}^2+F(\phi)\right)\mathrm{d}x= \int_\Omega \left(\frac{\varepsilon^2}{2}\abs{\nabla\phi}^2+\frac{1}{4}(\phi^2-1)^2\right)\mathrm{d}x,
     \end{equation}
     with $F(\phi)=\frac{1}{4}(\phi^2-1)^2$, and  the small parameter $\epsilon$ is the width of the interface between the two stable constant states $\phi=1$ and $\phi=-1$.

\section{Energy stable network (EStable-Net)}\label{energy_decay_formula}

In this section, we
present our energy stable neural network EStable-Net.
Based on the block architecture Autoflow of the DOSNet \cite{lan2023dosnet},
we 
enforce the  energy stable property along the network, which is consistent with the physical property of the gradient flow. This design is expected to provide a stable and interpretable network structure compared with a black-box neural network.

The network is under the supervised learning framework, which can be applied generally even when only date is available but the equation is unknown. The block network structure is to mimic the evolution with finite time increment. However, the condition of time step $\Delta t\rightarrow 0$ in many available neural network methods and the traditional numerical methods reviewed in the introduction for consistency and convergence is not required in our network, thus our network is much more efficient in inference than these methods once it is trained.

The block network structure Autoflow is in general convenient to incorporate  physical and/or numerical properties of the PDEs, not limited to the energy stable property of the EStable-Net presented in this paper.


%

\subsection{Network architecture: Autoflow}\label{sec:autoflow}

\begin{figure}[htbp]
  \centering
  \includegraphics[width=0.7\textwidth]{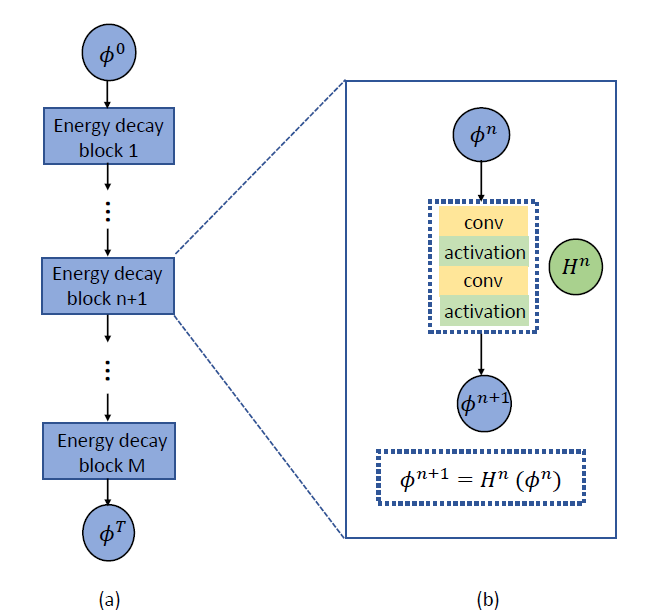}
  \caption{Network architecture of  EStable-Net. (a) The overall Autoflow structure of the network. (b) The structure within each energy stable block, which mimics the evolution from time $t_n$ to $t_{n+1}$, where $M$ is the number of blocks in the network. The input and output of each block are respectively $\phi^n$ and $\phi^{n+1}$, which mimic the intermediate states of the evolution process of the gradient flow equation with enforcement of the energy stable property. The neural network is trained via supervised learning at time $T$. } \label{fig..nn_structure}
\end{figure}

The network architecture that we use is a generalization of the block structure  Autoflow proposed in
 the DOSnet \cite{lan2023dosnet}, which has two levels of structures. On a coarse level, it is a general dynamical architecture Autoflow (autonomous flow) consisting of several learnable  blocks to mimic the action of the evolution operators; see Fig.~\ref{fig..nn_structure}a. Inside each block, 
 the structure is designed to incorporate the physical and/or numerical properties of the PDE, and here the design in each block is to enforce the energy stable property that is consistent with the gradient flow equations (whose formulation will be given in the next subsection); see the illustration in Fig.~\ref{fig..nn_structure}b.

\begin{figure}[htbp]
  \centering
  \includegraphics[width=\textwidth]{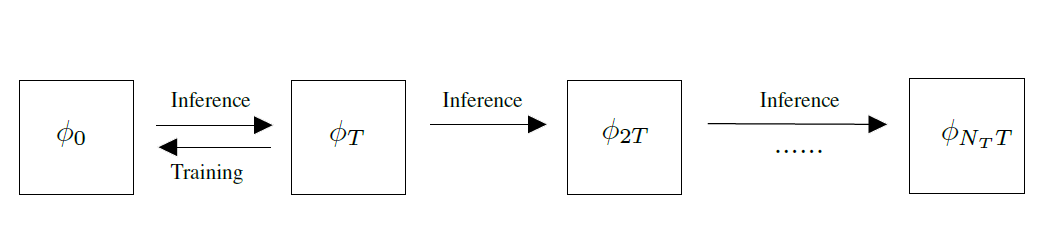}
  \caption{Inference of EStable-Net.} \label{fig..inference}
\end{figure}

Specifically, the architecture of the proposed energy stable neural networks EStable-Nets for solving gradient flow equations consists of a few energy stable blocks,  with input being the initial condition $\phi^0$ and the output the solution $\phi^T$ at time $T$. The output of the $n$-th block
includes the $\phi^n$, which have the same dimensions as the discretized solution $\phi$  and can be interpreted as the intermediate states of the evolution process of the gradient flow equation. The output of the $n$-th block also includes some auxiliary quantities to enforce the energy stable properties, which will be discussed in Sec.~\ref{subsec:arch}.
The output of the network is
\begin{equation} \label{eq:Autoflow}
	\mathcal{N}_{\pmb \theta}\left(\phi_0\right) = \psi_{ \theta_M}\circ\psi_{ \theta_{M-1}}\circ\cdots \circ \psi_{ \theta_n}\circ\cdots \circ \psi_{ \theta_1}\left(\phi_0\right),
\end{equation}
where $\psi_{\theta_n}$ is a neural network block with parameters $\theta_n$ and $M$ denotes the number of blocks inside Autoflow.

The neural network is trained via supervised learning at time $T$.
Using training data $(\phi^0, \phi^T)$,  e.g., generated by some high-precision classical numerical method, the learnable parameters $\boldsymbol{\theta} = \{\theta_n\}_{n = 1}^M$ inside energy decay network can be obtained via minimizing the mean square error loss (MSE):  $\boldsymbol{\theta}^* = \mathop{\arg\min}_{\pmb \theta}\mathcal{L}(\boldsymbol{\theta})$, with
\begin{equation}
    \mathcal{L}(\boldsymbol{\theta}) = \frac1N\sum_{i=1}^N\|\mathcal{N}_{\pmb \theta}(\phi^{(i)}(x,0))-\phi^{(i)}(x,T)\|^2
    +\mathcal{L}_{\Delta E}(\boldsymbol{\theta}), \label{eqn:loss}
\end{equation}
where $N$ is number of the training data, symbols with  superscript $i$ represents the $i$-th sample,   $\mathcal{N}_{\boldsymbol{\theta}}(\phi^{(i)}(x,0))$ is $i$-th neural network output and $\phi^{(i)}(x,T)$ is the ground truth, $\|\cdot\|$ is the $L^2$ norm, and the term
$\mathcal{L}_{\Delta E}(\boldsymbol{\theta})$ enforces the energy dissipation to be given in Eq.~\eqref{eqn:loss-diss} in the next subsection.

Note that the training time $T$ is a finite time, and the number of block $M$ is small, e.g. $M=5$, meaning that the equivalent time increment of each learning block $\Delta t=T/M$ is also finite. This is different from the classical numerical methods and most of the available neural network methods for solving evolution PDEs as reviewed in the introduction, in which it is required that the time step $\Delta t\rightarrow 0$ for consistency and convergence. Instead, in our neural network, the error is controlled by the supervised training at time $T$, and time step $\Delta t\rightarrow 0$ is not required.

The network is trained by using the evolution results at time $T$ from different initial values. Once the network is trained, it can be applied to infer the evolution results of the gradient flow equation at  time $t=2T, 3T, 4T,\cdots$,
or to infer the evolution from any initial value at  time $t=T, 2T, 3T, 4T,\cdots$; see Fig.~\ref{fig..inference}.
Stability of the neural network provides a basis for the inference for time matching.

\subsection{Energy stable blocks}\label{subsec:arch}

In our energy stable network EStable-Net, each block in the Autoflow network structure is designed to
incorporate the energy stable property.
The main input and output of each block   are respectively $\phi^n$ and $\phi^{n+1}$, which mimic the intermediate states of the evolution process of the gradient flow equation.
The formulation in each energy stable block is inspired by the  energy dissipation formulations in Eq.~\eqref{eqn:energy-diss} in Sec.~\ref{sec:gradient-flow}.

We focus on the Allen-Cahn equation and the Cahn-Hilliard equation in this paper. From Eq.~\eqref{eqn:e-ac-ch}, the energy of $\phi^k$ is
\begin{flalign}\label{eqn:discrte-e}
     E(\phi^k) = \int_\Omega \left(\frac{\varepsilon^2}{2}\abs{\nabla\phi^k}^2+F(\phi^k)\right)\mathrm{d}x.
\end{flalign}
The change of the discrete energy from $t_n$ to $t_{n+1}$ can be approximated by
\begin{flalign}
\delta E^n(\phi^n,\phi^{n+1})\approx \Delta E^n=E(\phi^{n+1})-E(\phi^n).
\end{flalign}
 For the Allen-Cahn equation, following Eq.~\eqref{eqn:energy-diss-ac}, we define
\begin{flalign}\label{eqn:delta-e-ac}
\delta E_{\rm ac}^n(\phi^n,\phi^{n+1})=- \frac{1}{\Delta t} \|\phi^{n+1}-\phi^n\|^2.
\end{flalign}
For the Cahn-Hilliard equation, following Eq.~\eqref{eqn:energy-diss-ch}, we define
\begin{flalign}\label{eqn:delta-e-ch}
\delta E_{\rm ch}^n(\phi^n,\phi^{n+1})= - \Delta t \left\|\nabla\frac{\delta E}{\delta \phi}(\phi^n)\right\|^2.
\end{flalign}

We define a loss that enforces the energy dissipation as
\begin{flalign}\label{eqn:loss-diss}
\mathcal{L}_{\Delta E}(\boldsymbol{\theta})=\beta\sum_{n=0}^{M-1} \Big[E(\phi^{n+1})-E(\phi^n)- \delta E^n(\phi^n,\phi^{n+1})\Big]^2,
\end{flalign}
where $\beta>0$ is some parameter for the incorporation of this energy dissipation loss into the total loss; see Eq.~\eqref{eqn:loss}. In the simulations of this paper, we choose $\beta=1$.
The integrals in this loss for $E^n$ and $\delta E(\phi^n)$ in Eqs.~\eqref{eqn:discrte-e} and \eqref{eqn:delta-e-ac}/\eqref{eqn:delta-e-ch}
are approximated by some summation scheme over the discrete points over the spatial domain, e.g. the trapezoidal rule for a uniform mesh. The gradient in these equations are calculated by the central difference scheme.

Note that unlike the classical numerical methods and most of the available neural network methods for solving evolution PDEs as reviewed in the introduction, in which it is required that the time step $\Delta t\rightarrow 0$ for consistency and convergence, in our Autoflow network structure and the EStable-Net, the error is controlled by the supervised training at time $T$, and time step $\Delta t\rightarrow 0$ is not required. The similarity between the update formulation in the EStable-Net and the discrete energy dissipation ensures that the evolution provided by the EStable-Net is not far from the actual gradient flow evolution path, which better incorporates the physics of the gradient flow equation compared with a general  black-box neural network approximation.

An alternative formulation that enforces decrease of a discrete version of the energy in an energy stable block is presented in Appendix~\ref{app:A}.

\section{Numerical experiments}\label{sec:numerical}
In this section, we perform numerical experiments of the EStable-Net on classical gradient flow equations of the Allen-Cahn equation in Eq.~\eqref{eqn:AC}
and the Cahn-Hilliard equation in Eq.~\eqref{eqn:CH}
in two dimensions.  The small parameter $\epsilon=0.02$. The spatial domain $\Omega=[-1,1]^2$ with periodic boundary conditions.

The spatial domain $\Omega=[-1,1]^2$ is discretized into $m\times m$ evenly spaced grid points ($128\times128$ for  the Allen-Cahn equation and $64\times64$ for  the Cahn-Hilliard equation). The inputs $\phi(\mathbf x,0)$ are generated by combing fourier basis function whose frequency less than $8$ and coefficients sampled from normal distribution.  The network is trained by the data of ground truth $\phi(\mathbf x,T)$ at time $T$ from given initial value $\phi(\mathbf x,0)$.  The label data $\phi(\mathbf x,T)$  is calculated from $\phi(\mathbf x,0)$ with high accuracy by the numerical schemes of pseudo-spectral method with the Fast Fourier Transform in space and ode45 solver in MATLAB is adopted for the time integration.

In the supervised training, we adopt the following strategy based on the physics of the gradient flow dynamics. In both the dynamics of the Allen-Cahn equation and the Cahn-Hilliard equation, two evolution stages can be identified, namely first the
phase separation stage in which the initial evolves into a binary system with the two energy minimum values $\phi=1$ and $\phi=-1$ with smoothed boundaries,  and then the coarsening stage in which the pattern characteristic length increases by disappearance of small patches or merge of small patches into bigger ones \cite{AC,CH1,CH2}.
We use both types of phase separation and coarsening date in the training, which has been shown to be able to significantly increase the accuracy of the network predictions in the training and the inference for later evolution, especially for the Cahn-Hilliard equation.

We choose the training time $T$ before which the phase separation stage ends and the coarsening stage starts.
The input-ground truth data for training include both the pairs $\big(\phi^{(i)}(\mathbf x,0),\phi^{(i)}(\mathbf x,T)\big)$ for the phase separation stage and the pairs $\big(\phi^{(i)}(\mathbf x,T),\phi^{(i)}(\mathbf x,2T)\big)$ for the coarsening stage. The label data $\phi(\mathbf x,2T)$  is also calculated from $\phi(\mathbf x,T)$ with high accuracy by the numerical schemes of pseudo-spectral method with the Fast Fourier Transform in space and ode45 solver in MATLAB for the time integration.

\subsection{Allen-Cahn equation in two dimensions}\label{sec:ac-2d}

In this subsection, we solve the Allen-Cahn equation in Eq.~\eqref{eqn:AC} in two dimensions.
We choose $M=5$ energy stable blocks in the network EStable-Net, and use two convolutional layers and Tanh activation function for each energy stable block.
We use the default (kaiming uniform) initialization for weights.
The optimizer of our model is Adam \cite{kingma2014adam} with initial learning rate $10^{-3}$ and $L^2$ regularization weight decay rate $10^{-7}$. The learning rate decays by half every 100 epochs, and restarts for each 1000 epochs.

The training time is $T=5$. We generate 5056 pairs of data $\big(\phi^{(i)}(\mathbf x,0),\phi^{(i)}(\mathbf x,5)\big)$ and $\big(\phi^{(i)}(\mathbf x,5),\phi^{(i)}(\mathbf x,10)\big)$ as described at the beginning of this section,  and 4096 of them are used for training while the left 960 are used for testing.

\begin{figure}
\centering
\subfigure[]{
\includegraphics[width=0.45\textwidth]{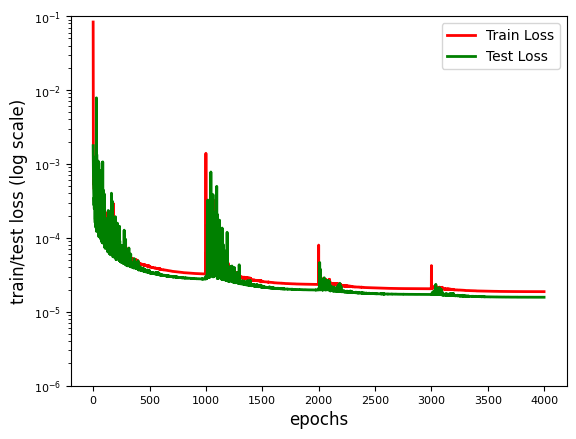}}
\subfigure[]{
\includegraphics[width=0.45\textwidth]{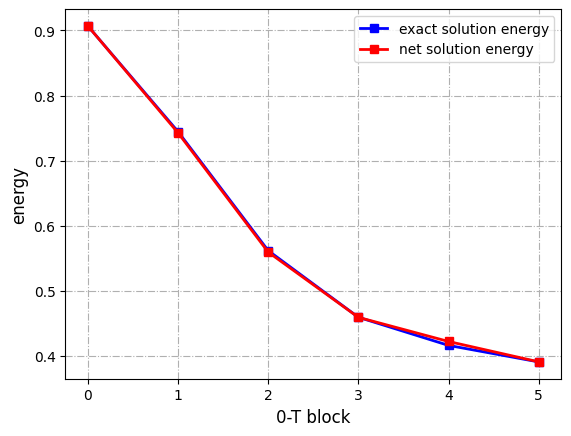}}
\subfigure[]{
\includegraphics[width=\textwidth]{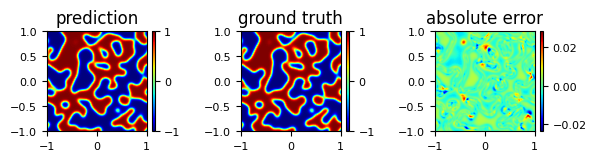}}
\caption{Allen-Cahn equation: (a) The train loss and the test loss (MSE). (b) Evolution of the energy of the EStable-Net solution and that of the exact solution.
(c) An example of the prediction of the EStable-Net  and the exact solution.}
\label{fig..AC2D_prediction}
\end{figure}

Results of accuracy of the EStable-Net on this Allen-Cahn equation are shown in Figure \ref{fig..AC2D_prediction}. After 4000 epochs of training, the train loss and test loss are reduced to about $1.57\times 10^{-5}$  as shown in Figure \ref{fig..AC2D_prediction}(a), and the relative $L^1$ error reaches $3.65\times10^{-3}$. Figure \ref{fig..AC2D_prediction}(c) shows an example, in which
the prediction of our network agrees well with the exact solution.

An example of evolution of the energy in Eq.~\eqref{eq..contenergy} of the trained EStable-Net and comparison of the exact energy evolution of the Allen-Cahn equation are shown in Figure \ref{fig..AC2D_prediction}(b). It can be seen from the figure that  the  energy given by EStable-Net agrees well with that of the exact solution, both are decreasing  block by block in the evolution.


\begin{figure}[htbp]
\centering
\includegraphics[width=0.8\textwidth]{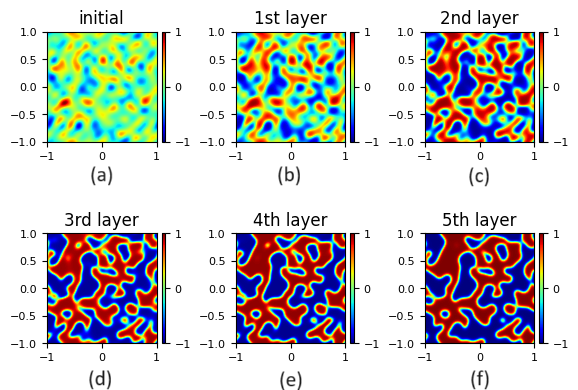}
\caption{Allen-Cahn equation: An example of the initial state and output of the 5 energy stable blocks in EStable-Net for the evolution from $0$ to $T=5$,  from  (a) to (f).}
\label{fig..AC2D_middle}
\end{figure}

\begin{figure}[htbp]
\centering
\includegraphics[width=0.8\textwidth]{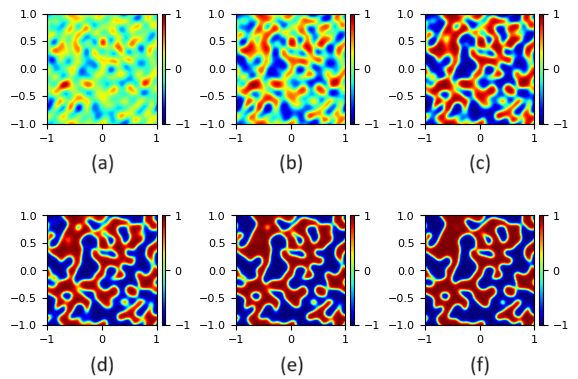}
\caption{Allen-Cahn equation: An example of snapshots in the evolution  with time increment $T/5=1$ obtained by numerical method using a much smaller time step, from (a)-(f). The initial condition is the same as that for the example in Fig.~\ref{fig..AC2D_middle}.}
\label{fig..AC2D_middlecheck}
\end{figure}

Figures \ref{fig..AC2D_middle} and \ref{fig..AC2D_middlecheck} show an example of the output of the 5 energy stable blocks in EStable-Net for the Allen-Cahn equation, and the corresponding snapshots in the evolution with equal time increment $T/5$  obtained by numerical method using a much smaller time step, respectively.
It can be seen that the evolution  given by the outputs of the  energy stable blocks in EStable-Net agrees well with the evolution of the solution of the Allen-Cahn equation.

\begin{table}[hbtp]
\centering
\begin{tabular}{lccccc}
\toprule
&   $\frac{1}{5} T$ & $\frac{2}{5} T$ & $\frac{3}{5} T$ & $\frac{5}{5} T$ & $T$ \\
\midrule
EStable-Net & 0.0131 & 0.0200 & 0.0303 & 0.0271 & 0.0037 \\
\midrule
block only & 2.9772 & 2.0220 & 1.7484& 1.7212& 0.0028\\
\bottomrule
\end{tabular}
\caption{Relative $L^1$ errors of (1) the EStable-Net and (2) the neural network without the energy stable design (the row of "block only"; see discussion in Section~\ref{sec:5.1}).}
\label{table1}
\end{table}

The errors of the predictions of EStable-Net  of the  intermediate and final blocks for the evolution from $0$ to $T$ are shown in Table~\ref{table1}. It can be seen that  EStable-Net is able to give accurate predictions for both final and intermediate stages during the evolution of the Allen-Cahn equation.

\begin{figure}[htbp]
\centering
\includegraphics[width=0.5\textwidth]{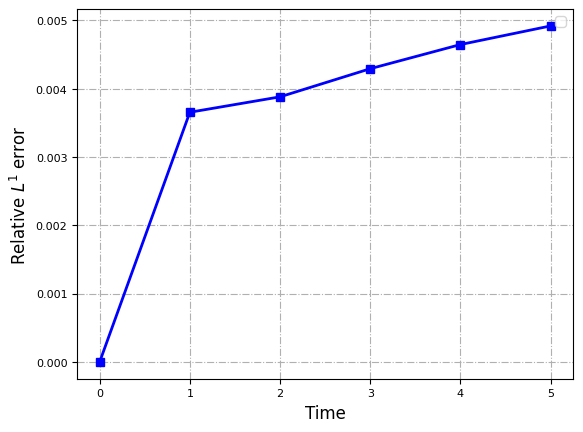}
\caption{Allen-Cahn equation: Relative $L^1$ errors at the training time $T$ and of the inference results at  time $t=2T, 3T, 4T,5T$. The time unit is $T=5$.}
\label{fig..ac-error}
\end{figure}


\begin{figure}[htbp]
\centering
\includegraphics[width=0.8\textwidth]{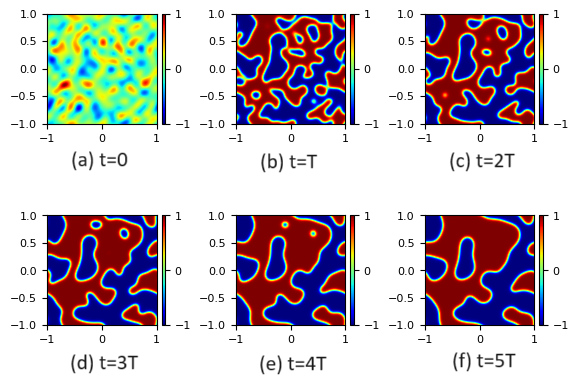}
\caption{Allen-Cahn equation:  An example of the initial state (a), the predicted state at the training time $t=T$ (b), and the inference results of the trained EStable-Net  at  time $t=2T, 3T, 4T,5T$  (c)--(f).}
\label{fig..AC2D_0-5}
\end{figure}

\begin{figure}[htbp]
\centering
\includegraphics[width=0.8\textwidth]{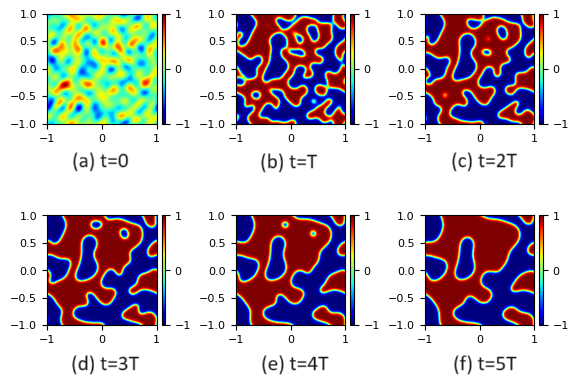}
\caption{Allen-Cahn equation:  An example of the initial state (a), snapshots in the evolution at time  $t=T, 2T, 3T, 4T,5T$ (b)--(f) obtained by numerical method. The initial condition is the same as that for the example in Fig.~\ref{fig..AC2D_0-5}.}
\label{fig..AC2D_exact0-5}
\end{figure}

We apply the trained network to infer the evolution results of the Allen-Cahn equation at  time $t=2T, 3T, 4T,5T$. The relative $L^1$ errors of these inference results are shown in Fig.~\ref{fig..ac-error}. It can be seen that the errors increase slowly in a linear manner from
$0.365\%$ at time $t=T$ to about $0.5\%$  at $t=5T$.
The average slope of the error increase of the network inference results from $T$ to $5T$ is smaller than the slope of the training error in $[0,T]$.
This behavior of errors can be understood by the fact that the training time interval $[0,T]$ includes both evolution stages of phase separation and coarsening, and the inference time periods $[T,2T], [2T,3T], [3T, 4T], [4T,5T]$ only involve the coarsening stage and the evolution is not as significant as that in the initial time interval $[0,T]$.

Figures \ref{fig..AC2D_0-5} and \ref{fig..AC2D_exact0-5} respectively show an example of the inference results of the trained EStable-Net for the Allen-Cahn equation at  time $t=2T, 3T, 4T,5T$ together with the initial state and the predicted state at the training time $t=T$,
and the corresponding snapshots in the evolution with same time increment obtained by numerical method.
It can be seen that the predictions of the EStable-Net agree excellently with the evolution of the solution the Allen-Cahn equation.

\subsection{Cahn-Hilliard equation in two dimensions}

In this subsection, we solve the Allen-Cahn equation in Eq.~\eqref{eqn:CH} in two dimensions.
We choose $M=4$ energy stable blocks in the network EStable-Net, and use two convolutional layers and Tanh activation function  in each energy stable block.
We use the default (kaiming uniform) initialization for weights.
The optimizer of our model is Adam \cite{kingma2014adam} with initial learning rate $10^{-3}$ and $L^2$ regularization weight decay rate $10^{-7}$. The learning rate decays by half every 100 epochs, and restarts for each 1000 epochs.

The training time is $T=0.2$. We generate 5120 pairs of data $\big(\phi^{(i)}(\mathbf x,0),\phi^{(i)}(\mathbf x,5)\big)$ and $\big(\phi^{(i)}(\mathbf x,5),\phi^{(i)}(\mathbf x,10)\big)$ as described at the beginning of this section,  and 4096 of them are used for training while the left 1024 are used for testing.

\begin{figure}[htbp]
\centering
\subfigure[]{
\includegraphics[width=0.45\textwidth]{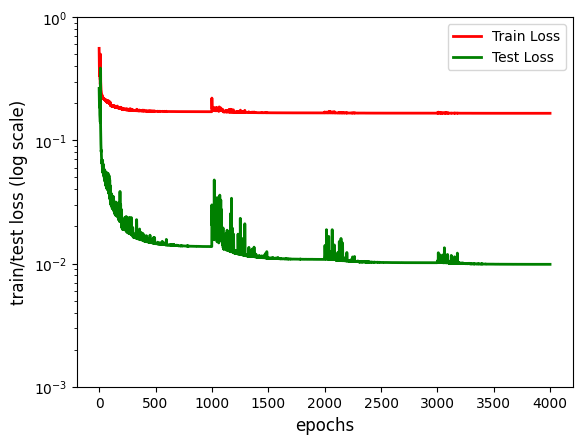}}
\subfigure[]{
\includegraphics[width=0.45\textwidth]{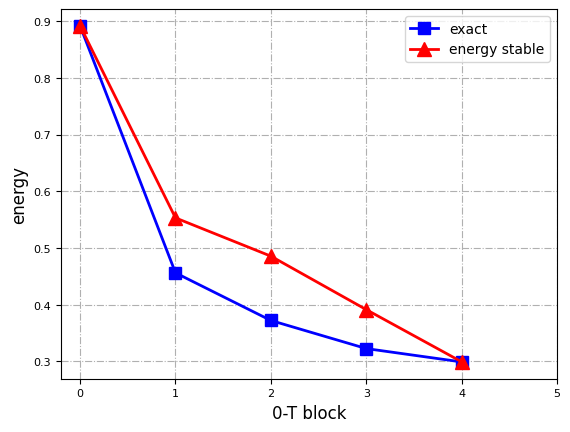}}
\subfigure[]
{\includegraphics[width=\textwidth]{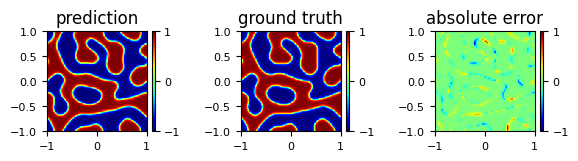}}
\caption{Cahn-Hilliard equation:  (a) The total train loss and the test loss. Note that the MSE train loss is similar to that of the test loss. (b) Evolutions of the energy of the EStable-Net solution and that of the exact energy. (c) An example of
the prediction of our network and the exact solution.}
\label{fig..CH2D_prediction}
\end{figure}

Results of accuracy of the EStable-Net on this Cahn-Hilliard equation are shown in Figure \ref{fig..CH2D_prediction}(a). After 4000 epochs of training, the MSE train loss and test loss are reduced to about $9.85\times 10^{-3}$   as shown in Figure \ref{fig..CH2D_prediction}(a), and the relative $L^1$ error reaches $0.045$.
Note that in Figure \ref{fig..CH2D_prediction}(a), the total training loss is relatively large because it also includes the loss of energy dissipation, and the behavior of the MSE train loss is similar to that of the test loss.
Figure \ref{fig..CH2D_prediction}(c) shows an example, in which
the prediction of our network agrees well with the exact solution.

An example of evolution of the  energy in Eq.~\eqref{eq..contenergy} of the trained EStable-Net and comparison of the exact energy evolution of the Cahn-Hilliard equation are shown in Figure \ref{fig..CH2D_prediction}(b). It can be seen from the figure that  the energy given by the EStable-Net solution is decreasing block by block, and the decrease is consistent with that of the energy of the exact solution.

\begin{figure}[htbp]
\centering
\includegraphics[width=\textwidth]{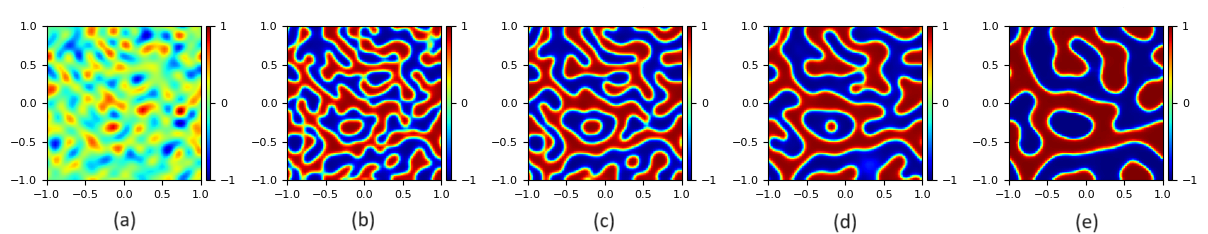}
\caption{Cahn-Hilliard equation: An example of the initial state and output of the 4 energy stable blocks in EStable-Net for the evolution from $0$ to $T=0.2$,  from  (a) to (e).}
\label{fig..CH2D_middle}
\end{figure}

\begin{figure}[htbp]
\centering
\includegraphics[width=\textwidth]{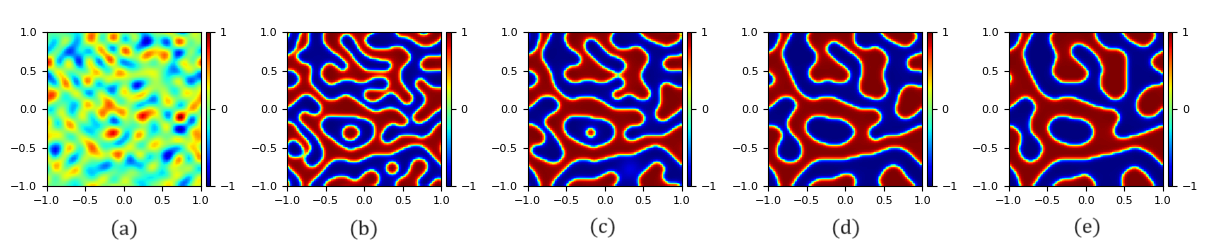}
\caption{Cahn-Hilliard equation: An example of snapshots in the evolution  with time increment $T/4=0.05$ obtained by numerical method using a much smaller time step, from (a)-(e). The initial condition is the same as that for the example in Fig.~\ref{fig..CH2D_middle}.}
\label{fig..CH2D_middlecheck}
\end{figure}

Figures \ref{fig..CH2D_middle} and \ref{fig..CH2D_middlecheck} show an example of the output of the 4 energy stable blocks in EStable-Net for the Cahn-Hilliard equation, and the corresponding snapshots in the evolution with equal time increment $T/4$  obtained by numerical method using a much smaller time step, respectively.
It can be seen that the evolution trend given by the output of the  energy stable blocks in EStable-Net essentially agrees with the evolution of the solution of Cahn-Hilliard equation.

\begin{figure}[htbp]
\centering
\includegraphics[width=0.5\textwidth]{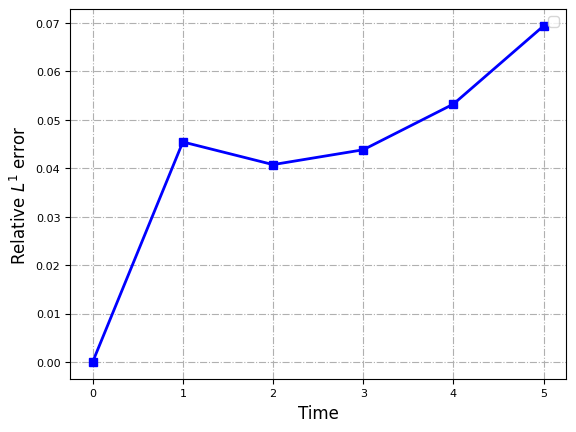}
\caption{Cahn-Hilliard equation: Relative $L^1$ errors at the training time $T$ and of the inference results at  time $t=2T, 3T, 4T,5T$. The time unit is $T=0.2$.}
\label{fig..ch-error}
\end{figure}

We apply the trained network to infer the evolution results of the Cahn-Hilliard equation at  time $t=2T, 3T, 4T,5T$. The relative $L^1$ errors of these inference results are shown in Fig.~\ref{fig..ch-error}. It can be seen the prediction errors increase slowly with  time,  from $4.5\%$ at $t=T$ to about $7\%$ at $t=5T$, except at $t=2T$ where the energy is slightly decreased.
The average slope of the error increase of the network inference results from $T$ to $5T$ is smaller than the slope of the training error in $[0,T]$. As in the case of the Allen-Cahn equation, this behavior of errors can be understood by the fact that the training time interval $[0,T]$ includes both evolution stages of phase separation and coarsening, and the inference time periods $[T,2T], [2T,3T], [3T, 4T], [4T,5T]$ only involve the coarsening stage and the evolution is not as significant as that in the initial time interval $[0,T]$.

\begin{figure}[htbp]
\centering
\includegraphics[width=0.8\textwidth]{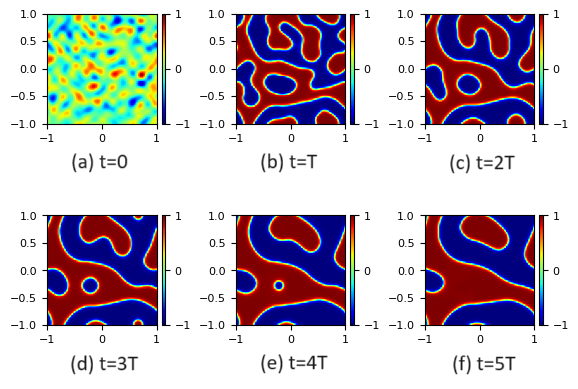}
\caption{Cahn-Hilliard equation: An example of the initial state (a), the predicted state at the training time $t=T$ (b), and the inference results of the trained EStable-Net  at  time $t=2T, 3T, 4T,5T$  (c)--(f). }
\label{fig..CH2D_0-5}
\end{figure}

\begin{figure}[htbp]
\centering
\includegraphics[width=0.8\textwidth]{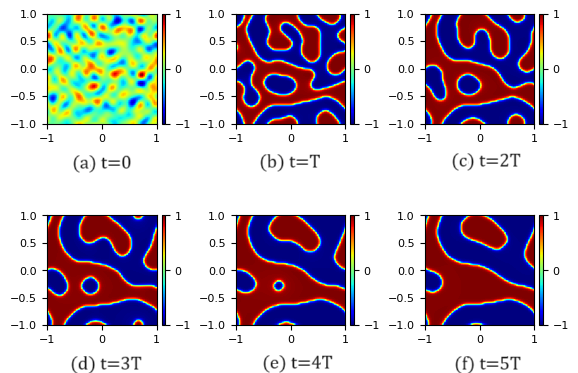}
\caption{Cahn-Hilliard equation: An example of the initial state (a), snapshots in the evolution at time  $t=T, 2T, 3T, 4T,5T$ (b)--(f) obtained by numerical method. The initial condition is the same as that for the example in Fig.~\ref{fig..CH2D_0-5}.}
\label{fig..CH2D_exact0-5}
\end{figure}

Figures \ref{fig..CH2D_0-5} and \ref{fig..CH2D_exact0-5} respectively show an example of the inference results of the trained EStable-Net for the Cahn-Hilliard equation at  time $t=2T, 3T, 4T,5T$ together with the initial state and the predicted state at the training time $t=T$,
and the corresponding snapshots in the evolution with same time increment obtained by numerical method.
It can be seen that the predictions of the EStable-Net agree excellently with the evolution of the Cahn-Hilliard equation.

\section{Discussion}\label{sec:dis}

In this section, we further examine the energy stable network design and the physics based two-stage training strategy.

\subsection{With/Without energy stable design in the neural network}\label{sec:5.1}

\begin{figure}[htpb]
\centering
\subfigure[]{
\includegraphics[width=0.45\textwidth]{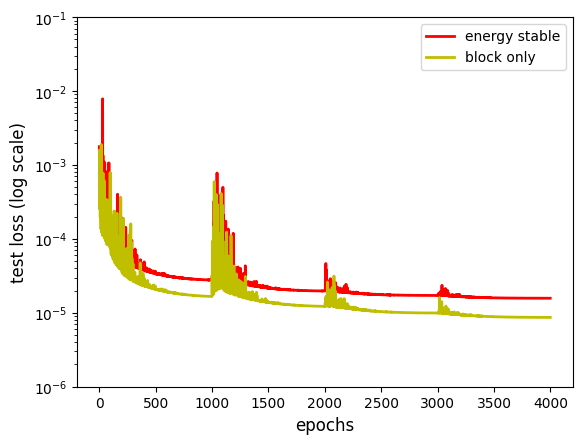}}
\subfigure[]{
\includegraphics[width=0.45\textwidth]{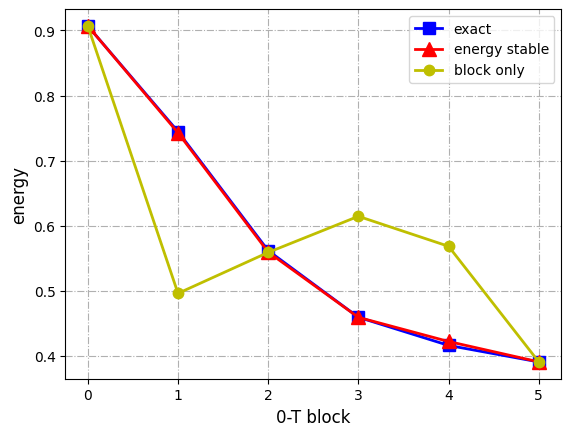}}
\caption{Allen-Cahn equation:  With/Without energy stable design. (a) The train loss and the test loss (MSE). (b) Evolution of the energy of the EStable-Net solution and that of the exact solution. The results of "block only" are those of neural network without energy stable design.}
\label{fig..AC2D_no-stable}
\end{figure}

We compare the results of the neural network with and without the energy stable design for solving the Allen-Cahn equation under the same settings as in Section~\ref{sec:ac-2d}. The comparison results of accuracy and energy evolution are shown in Figure~\ref{fig..AC2D_no-stable}. As shown in Figure~\ref{fig..AC2D_no-stable}(b), the  energy given by EStable-Net agrees well with that of the exact solution, and both are decreasing  block by block in the evolution (already shown in Figure \ref{fig..AC2D_prediction}(b) previously); whereas without the energy stable design, the energy of the prediction of the neural network is not monotonic and does not agree with the exact energy evolution. The train loss and test loss of the EStable-Net are slightly higher than those of the network without the energy stable design as shown by Figure~\ref{fig..AC2D_no-stable}(a), which can be understood as the small cost to enforce the energy stable property.

\begin{figure}[htbp]
\centering
\includegraphics[width=0.8\textwidth]{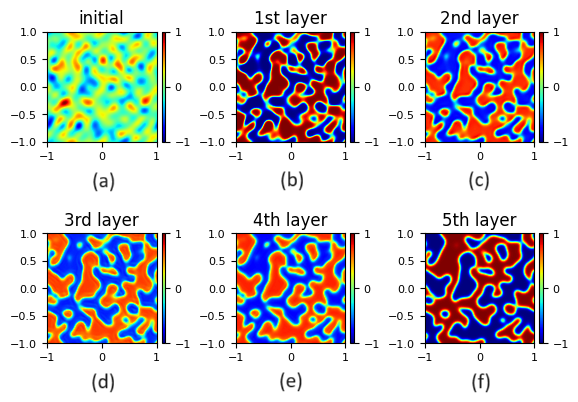}
\caption{Allen-Cahn equation: An example of the initial state and output of the 5 blocks without energy stable design for the evolution from $0$ to $T=5$,  from  (a) to (f).}
\label{fig..AC2D_middle no-stable}
\end{figure}

Figure~\ref{fig..AC2D_middle no-stable} show an example of the output of the 5 blocks without energy stable design for the Allen-Cahn equation from $0$ to $T=5$. Other than the energy stable design in the neural network, the initial condition and other neural network settings are the same as the EStable-Net whose results are given in Figure \ref{fig..AC2D_middle}. Comparing the results in Figure~\ref{fig..AC2D_middle no-stable} with the EStable-Net results in Figure \ref{fig..AC2D_middle} and the results of numerical method using a much smaller time step shown in Figure  \ref{fig..AC2D_middlecheck}, 
it can be seen that the intermediate results in the evolution predicted by the  neural network without energy stable design do not agree  with the evolution of the solution of the Allen-Cahn equation, unlike the excellent agreement between the  EStable-Net results and the numerical results. It can also be seen from Table~\ref{table1} in the previous section that the errors of the predictions of the neural network without the energy stable design  for the  intermediate stages of the evolution are large,  unlike the accurate predictions of the intermediate stages of the evolution predicted by  the  EStable-Net.

\subsection{Training with/without data of $[T,2T]$}

\begin{figure}[htbp]
\centering
\subfigure[]{
\includegraphics[width=0.45\textwidth]{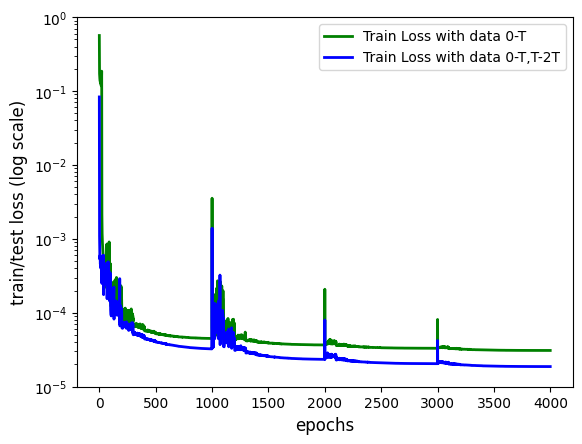}}
\subfigure[]{
\includegraphics[width=0.45\textwidth]{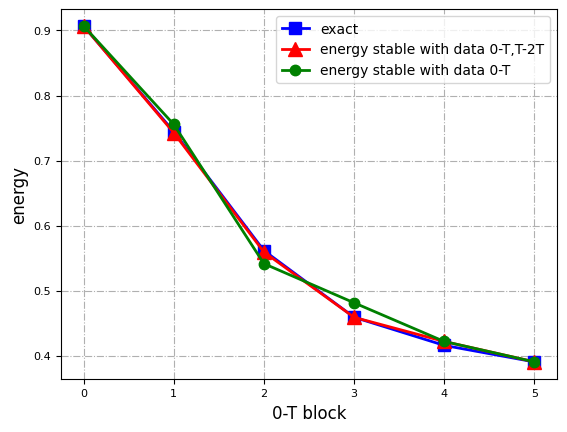}}
\subfigure[]{
\includegraphics[width=0.45\textwidth]{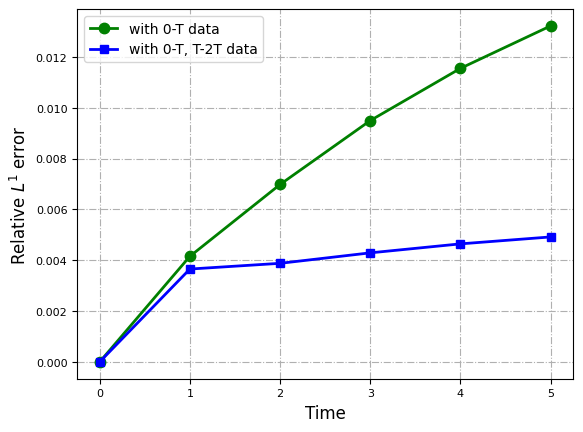}}
\caption{Allen-Cahn equation: Training with/without data of $[T,2T]$. (a) The Training loss. (b) Evolution of the energy. 
(c) Errors at the training time $T$ and of the inference results at time $t=2T,3T,4T,5T$.}
\label{fig..AC2D_T-2T}
\end{figure}

In the experimental results presented in the previous section, the training is based on the data of evolution of both time intervals $[0,T]$ and $[T,2T]$ to incorporate the physics of both the phase separation stage and the coarsening stage of the Allen-Cahn equation and the Cahn-Hilliard equation. Now we examine the effectiveness of this strategy by removing the data of $[T,2T]$ for training the EStable-Net for the Allen-Cahn equation. Comparisons of the results with and without the data of $[T,2T]$ are shown in  Figure~\ref{fig..AC2D_T-2T}. It can be seen that without the training data of $[T,2T]$, the training errors and the errors of the energy evolution for the time interval $[0,T]$ are slightly increased, while the errors of the inference at time $t=2T,3T,4T,5T$ are significantly increased. These demonstrate the effectiveness of incorporation of the training data of $[T,2T]$.

\section{Conclusions}\label{sec:conclusion}

   In this paper, we propose an 
    energy stable neural network (EStable-Net) for solving gradient flow equations. We first rewrite the gradient flow equation into an equivalent form by introducing an auxiliary variable.
     The solution update scheme in our neural network EStable-Net is inspired by this  auxiliary variable based equivalent form of the gradient flow equation, and enables decreasing of a discrete energy along the neural network.

     The architecture of the neural network EStable-Net consists of a few energy decay blocks, and the output of each block can be interpreted as an intermediate state of the evolution process of the gradient flow equation, following the Autoflow structure in the neural network DOSnet \cite{lan2023dosnet}. This design of identical input and output dimensions in the intermediate blocks  vastly reduces the parameters of the neural network.
          We prove that an discrete energy calculated
after each block is decreasing along the network,  which is consistent with the property in the evolution process of the gradient flow equation.
The proposed EStable-Net  can be applied generally to a wide range of gradient flow equations.

 We perform numerical experiments of the EStable-Net on  the Allen-Cahn equation in one and two dimensions. The numerical experimental results demonstrate that our network is able to generate  high accuracy and stable predictions.



\section*{Acknowledgments}
The work of Y Xiang was supported in part by HKUST IEG19SC04 and the Project of Hetao Shenzhen-HKUST Innovation Cooperation Zone HZQB-KCZYB-2020083. The work of LC Zhang was supported by National Natural Science Foundation of China 12201423 and
Shenzhen Science $\&$ Technology Innovation Program RCYX20231211090222026.

\appendix
\numberwithin{equation}{section}
\setcounter{equation}{0}
\numberwithin{theorem}{section}
\setcounter{theorem}{0}
\numberwithin{figure}{section}
\setcounter{figure}{0}

\section{An alternative formulation of the energy stable network}\label{app:A}

\subsection{Auxiliary variables and equivalence system of the gradient flow equation}\label{subsec:aux}

For the gradient flow equation \eqref{eq..gradflow},
we introduce  auxiliary variables  $U$ and $V$ as
\begin{flalign}
U=&\sqrt{\frac{k}{2}\abs{\mathcal{G}^{-1/2}\phi}^2+\frac12 \abs{\mathcal{D}^{1/2}\phi}^2+F(\phi)}, \vspace{1ex}  \label{eq..u}\\
V=&\sqrt{k}\mathcal{G}^{-1/2}\phi,  \label{eq..v}
\end{flalign}
where  $k$ is some positive constant.
The energy in Eq.~\eqref{eq..contenergy} can be written as
\begin{equation}\label{eqn:newenergy}
E=\|U\|^2-\frac{1}{2}\|V\|^2.
\end{equation}

The gradient flow equation in Eq.~\eqref{eq..gradflow}, using variables $U$ and $\phi$, can be written as
\begin{flalign}
\phi_t =&k\phi -\mathcal{G}(J_1(\phi)U),\label{eq..modif1}\\
U_t=&\frac{1}{2}J_2(\phi)\phi_t,\label{eq..modif2}\\
V=&\sqrt{k}\mathcal{G}^{-1/2}\phi,\label{eq..modif3}
\end{flalign}
where
\begin{flalign}
    J_1(\phi) =&\frac{k\mathcal{G}^{-1}\phi+\mathcal{D}\phi+f(\phi)}{\sqrt{\frac{k}{2} \abs{\mathcal{G}^{-1/2}\phi}^2+\frac12 \abs{\mathcal{D}^{1/2}\phi}^2+F(\phi)}}, \label{eq..modif4-1}\\
    J_2(\phi) = &\frac{k \mathcal{G}^{-1/2}\phi \mathcal{G}^{-1/2}+\mathcal{D}^{1/2}\phi \mathcal{D}^{1/2}+f(\phi)}{\sqrt{\frac{k}{2} \abs{\mathcal{G}^{-1/2}\phi}^2+\frac12 \abs{\mathcal{D}^{1/2}\phi}^2+F(\phi)}},\label{eq..modif4-2}
\end{flalign}
with the initial conditions
\begin{flalign}
    \phi(0,x) &= \phi_0(x),\label{eq..modif5}\\
    U(0,x) &= \sqrt{\frac{k}{2} \abs{\mathcal{G}^{-1/2}\phi_0}^2+\frac12 \abs{\mathcal{D}^{1/2}\phi_0}^2+F(\phi_0)}.\label{eq..modif6}\\
    V(0,x) &= \sqrt{k}\mathcal{G}^{-1/2}\phi_0.\label{eq..modif7}
\end{flalign}
It is easy to see that the new system \eqref{eq..modif1}-\eqref{eq..modif7} is equivalent to the original gradient flow system \eqref{eq..gradflow}-\eqref{eq..gradflowi}.

    The idea of introducing auxiliary variable and reformulating the original system into an equivalent system has been used in  \cite{yang2016linear,yang2017numerical,yang2017numerical2,pan2023novel}, where the purpose of the variable substitution is to reformulate the nonlinear terms in the free energy into quadratic form,  leading to a linear numerical scheme and in the meantime, maintaining the energy dissipation property.
 Here, we follow similar idea but use different auxiliary variables as in Eqs.~\eqref{eq..u} and \eqref{eq..v} with the new energy form in  Eq.~\eqref{eqn:newenergy}.

\subsection{Alternative energy stable blocks}\label{subsec:arch-A}

Our alternative formulation of  each energy stable block in the Autoflow network structure in Sec.~\ref{sec:autoflow}
is inspired by the new energy form in  Eq.~\eqref{eqn:newenergy} and the new evolution system in Eqs.~\eqref{eq..modif1}-\eqref{eq..modif3}.

The update formulation within each energy stable block is
\begin{flalign}
U^{n+1}=&\frac{1}{2}H^n(\phi^n)V^n\label{eqn:stable1}\\
V^{n+1}=&H^n(\phi^n)\left(U^{n+1}-U^{n} \right)+V^{n}\label{eqn:stable2}\\
\phi^{n+1}=&F^n(V^{n+1}).\label{eqn:stable3}
\end{flalign}
Here $H^n$ and $F^n$ are two neural networks, in which the $\tanh(\cdot)$ function is adopted as the activation function.   The main input and output of this block are respectively $\phi^n$ and $\phi^{n+1}$, which mimic the intermediate states of the evolution process of the gradient flow equation. The values $U^n$, $V^n$ and $U^{n+1}$, $V^{n+1}$ are the input and output values of the auxiliary variables $U$ and $V$  at time $t_n$ and $t_{n+1}$ to ensure the energy stable property. The initial value $V^0$ is learnt from $\phi^0$. 
Note that the boundness of the intermediate states $\phi^n$ and the final output $\phi^M$ is easily enforced by the $\tanh(\cdot)$ activation function used in the networks $F^n$ in Eq.~\eqref{eqn:stable3}.

We define the discrete energy
\begin{equation}\label{eqn:denergy}
E(U^{n},V^{n})=\|U^{n}\|^2-\frac{1}{2}\|V^{n}\|^2.
\end{equation}
Note that in this discrete energy, both $U^{n}$ and $V^{n}$ depend on $\phi^{n}$ in an implicit way.
 We have the following energy stable property for this discrete energy.

\begin{theorem}[Energy stable property]\label{thm1}
\begin{equation}\label{eqn:dissipation}
E(U^{n+1},V^{n+1})=E(U^{n},V^{n})-\|U^{n+1}-U^{n}\|^2-\frac{1}{2}\|V^{n+1}-V^{n}\|^2.
\end{equation}
\end{theorem}

\begin{proof}
Using Eqs.~\eqref{eqn:stable1} and \eqref{eqn:stable2}, we have
\begin{equation}
2U^{n+1}\left(U^{n+1}-U^{n}\right)=V^{n}\left(V^{n+1}-V^{n}\right).
\end{equation}
Further using $2a(a-b)=a^2+(a-b)^2-b^2$, we have
\begin{flalign}
\|U^{n+1}\|^2-\frac{1}{2}\|V^{n+1}\|^2=&\|U^{n}\|^2-\frac{1}{2}\|V^{n}\|^2-\|U^{n+1}-U^{n}\|^2-\frac{1}{2}\|V^{n+1}-V^{n}\|^2.
\end{flalign}
From the definition of the discrete energy in Eq.~\eqref{eqn:denergy},
 we have the energy stable property in Eq.~\eqref{eqn:dissipation}. Especially,
\begin{equation}
E(U^{n+1},V^{n+1})\leq E(U^{n},V^{n}).
\end{equation}
\end{proof}

In this sense, the update formulation Eqs.~\eqref{eqn:stable1}-\eqref{eqn:stable3} leads to an energy stable block.
Recall that $U^{n}$ and $V^{n}$ as well as the discrete energy in Eq.~\eqref{eqn:denergy} are determined by the supervised learning with loss at time $T$. This discrete energy is not necessarily the original energy in Eq.~\eqref{eq..contenergy}.
The update formulation in Eqs.~\eqref{eqn:stable1}-\eqref{eqn:stable3} and the associated energy stable property can be applied generally when only date is available without knowing the evolution equation.

We also have the following property for this alternative formulation in an energy stable block, which imposes a control on the discrete energy and the update formulation in Eqs.~\eqref{eqn:stable1}-\eqref{eqn:stable3}.
\begin{theorem}[Boundness of $\{V^{n}\}$]
Under the assumption $H^n=H$ for all $n$, we have that $\{V^{n}\}$ is bounded.
\end{theorem}

\begin{proof}
From the update formulation in Eqs.~\eqref{eqn:stable1} and \eqref{eqn:stable2}, we have
$$ V^{n+1}=H^n\left(\frac{1}{2}H^nV^n-\frac{1}{2}H^{n-1}V^{n-1} \right)+V^{n}.$$
Under the assumption $H^n=H^{n-1}=H$, we have
$$ V^{n+1}=\left(\frac{1}{2}H^2+1\right)V^n-\frac{1}{2}H^2V^{n-1}.$$
The general solution is
$$ V^n=C_1\left(\frac{1}{2}H^2\right)^n+C_2,$$
where constants $C_1$ and $C_2$ are determined by $V^0$ and $V^1$.
Since the absolute value $H^n$ is bounded by $1$ due to the activation function $\tanh(\cdot)$, we have that $\{V^{n}\}$ is bounded. \end{proof}

The alternative formulation of the energy stable network can be considered as a generalization of DOSnet \cite{lan2023dosnet} in the sense that it is based on a general energy splitting  through auxiliary variables for a gradient flow equation and is not based on the exact solution of any part in the splitting form. 

This energy stable network update formulation in Eqs.~\eqref{eqn:stable1}-\eqref{eqn:stable3}  is inspired by a numerical discretization of the auxiliary variable-based form of the evolution equation. That is, if we discretize  Eqs.~\eqref{eq..modif1}-\eqref{eq..modif3} with $k=\frac{1}{\Delta t}$, we have a similar system as Eqs.~\eqref{eqn:stable1}-\eqref{eqn:stable3},  with the expected approximation $F^n=\sqrt{\Delta t}\mathcal{G}^{1/2}$ and accordingly
$J_2(\phi)\phi_t \approx \frac{\frac{1}{\Delta t} \mathcal{G}^{-1/2}\phi \mathcal{G}^{-1/2}\phi_t}{U}=\frac{VV_t}{U}$.
Note again that the error of this network is controlled by the supervised training at time $T$, and time step $\Delta t\rightarrow 0$ is not required.

\subsection{Numerical experiments using the alternative formulation: Allen-Cahn equation}

In this subsection, we solve the Allen-Cahn equation in Eq.~\eqref{eqn:AC} in two dimensions using the alternative energy stable network presented above.
The problem and the network settings, except the energy stable blocks, are the same as those in the simulations in Sec.~\ref{sec:ac-2d}. Simulation results of the Cahn-Hilliard equation in two dimensions using the alternative energy stable network have similar behaviors in the convergence and accuracy, with slightly higher errors.

%
%
%

\begin{figure}[htbp]
\centering
\subfigure[]{
\includegraphics[width=0.45\textwidth]{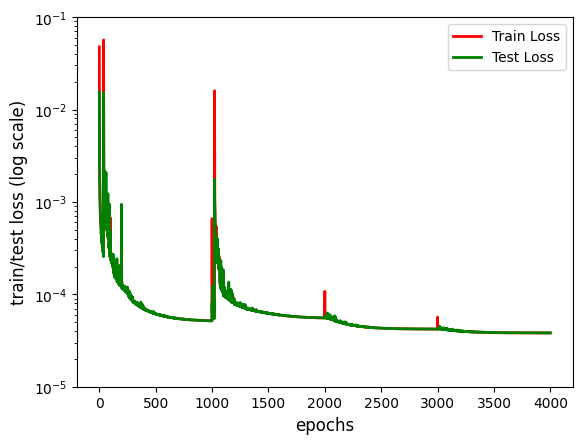}}
\subfigure[]
{\includegraphics[width=0.45\textwidth]{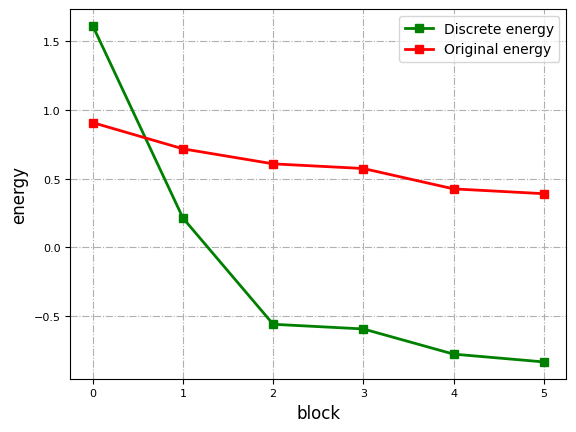}}
\subfigure[]
{\includegraphics[width=\textwidth]{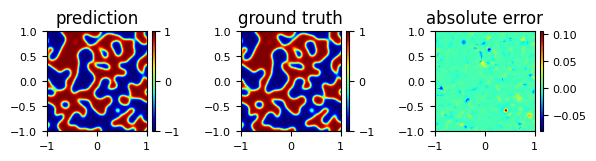}}
\caption{Allen-Cahn equation solved using the alternative energy stable network. (a) The train loss and the test loss (MSE). (b) Evolutions of the discrete energy in the network and the exact energy. (c) An example of
the prediction of the network and the exact solution. }
\label{fig..AC2D_prediction-A}
\end{figure}

Results of accuracy of this network on the Allen-Cahn equation are shown in Figure \ref{fig..AC2D_prediction-A}(a). After 4000 epochs of training, the train loss and test loss are reduced to about $3.8\times 10^{-5}$, and the relative $L^1$ error reaches $3.8\times10^{-3}$. Figure \ref{fig..AC2D_prediction-A}(c) shows an example, in which
the prediction of our network agrees well with the exact solution.

An example of evolution of the discrete energy in Eq.~\eqref{eqn:denergy} of the trained network and comparison of the exact energy evolution of the Allen-Cahn equation are shown in Figure \ref{fig..AC2D_prediction-A}(b). It can be seen from the figure that  the  discrete energy is indeed decreasing block by block, as predicted by analysis, which is consistent with the behavior of the exact energy of the Allen-Cahn equation.

\begin{figure}[htbp]
\centering
\includegraphics[width=0.8\textwidth]{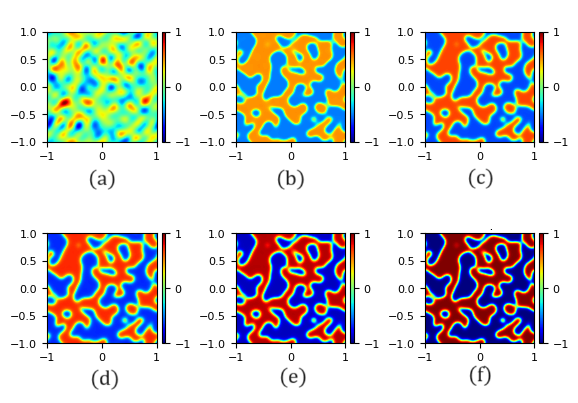}
\caption{Allen-Cahn equation solved using the alternative energy stable network: An example of the initial state and output of the 5 energy stable blocks in the network for the evolution from $0$ to $T=5$,  from  (a) to (f).}
\label{fig..AC2D_middle-A}
\end{figure}

\begin{figure}[htbp]
\centering
\includegraphics[width=0.8\textwidth]{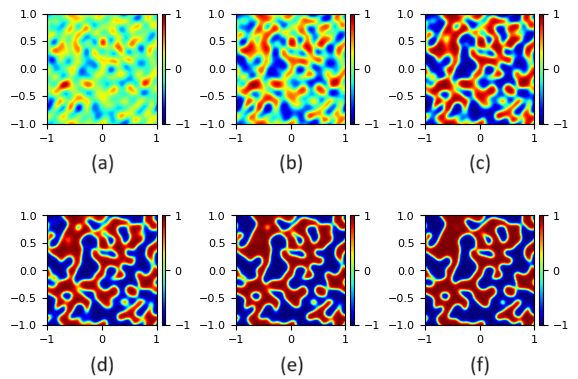}
\caption{Allen-Cahn equation: An example of snapshots in the evolution  with time increment $T/5=1$ obtained by numerical method, from (a)-(f). The initial condition is the same as that for the example in Fig.~\ref{fig..AC2D_middle-A}.}
\label{fig..AC2D_middlecheck-A}
\end{figure}

Figures \ref{fig..AC2D_middle-A} and \ref{fig..AC2D_middlecheck-A} show an example of the output of the 5 energy stable blocks in the network for the Allen-Cahn equation, and the corresponding snapshots in the evolution with equal time increment $T/5$  obtained by numerical method, respectively.
It can be seen that the evolution trend given by the outputs of the  energy stable blocks in the network are consistent with the evolution of the Allen-Cahn equation. 

\begin{figure}[htbp]
\centering
\includegraphics[width=0.45\textwidth]{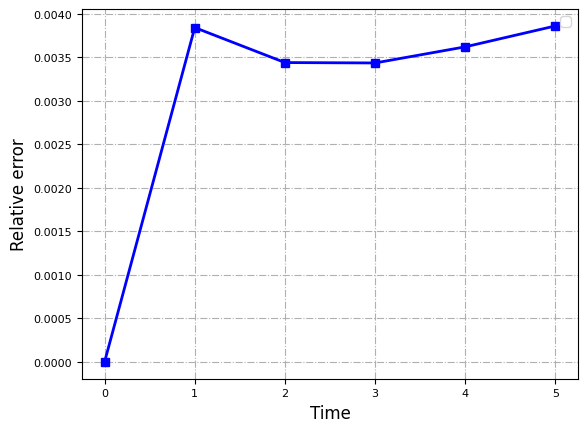}
\caption{Allen-Cahn equation solved using the alternative energy stable network:  $L^1$ errors at the training time $T$ and of the inference results at  time $t=2T, 3T, 4T,5T$. The time unit is $T=5$.}
\label{fig..ac-error-A}
\end{figure}

We apply the trained network to infer the evolution results of the Allen-Cahn equation at  time $t=2T, 3T, 4T,5T$. The relative $L^1$ errors of these inference results are shown in Fig.~\ref{fig..ac-error-A}. It can be seen the errors remain on the same level as the training error, around $0.4\%$, for these $4$ inference evolution results, and start to get bigger at $t=5T$. This behavior of errors can be understood by the fact that the training time interval $[0,T]$ includes both evolution stages of phase separation and coarsening.
Note that in the inference of the Cahn-Hilliard equation using the alternative energy stable network, the error is linearly increasing.

\begin{figure}[htbp]
\centering
\includegraphics[width=0.8\textwidth]{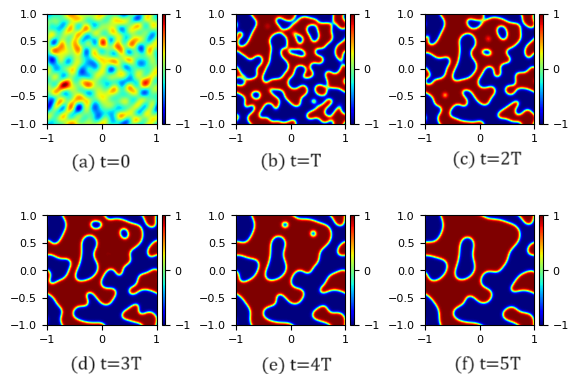}
\caption{Allen-Cahn equation:  An example of the initial state (a), the predicted state at the training time $t=T$ (b), and the inference results of the trained EStable-Net  at  time $t=2T, 3T, 4T,5T$  (c)--(f).}
\label{fig..AC2D_0-5}
\end{figure}

\begin{figure}[htbp]
\centering
\includegraphics[width=0.8\textwidth]{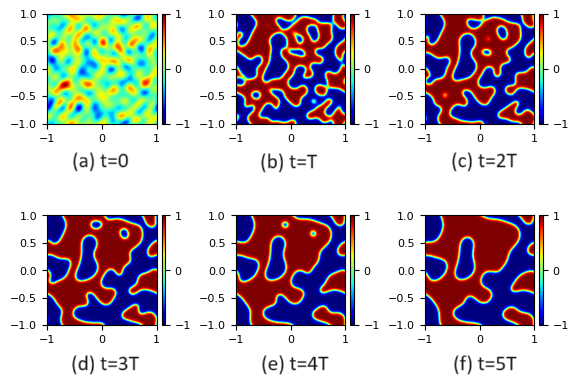}
\caption{Allen-Cahn equation:  An example of the initial state (a), snapshots in the evolution at time  $t=T, 2T, 3T, 4T,5T$ (b)--(f) obtained by numerical method. The initial condition is the same as that for the example in Fig.~\ref{fig..AC2D_0-5}.}
\label{fig..AC2D_exact0-5}
\end{figure}


Figures \ref{fig..AC2D_0-5} and \ref{fig..AC2D_exact0-5} respectively show an example of the inference results of the trained EStable-Net for the Allen-Cahn equation at  time $t=2T, 3T, 4T,5T$ together with the initial state and the predicted state at the training time $t=T$,
and the corresponding snapshots in the evolution with same time increment obtained by numerical method.
It can be seen that the predictions of the EStable-Net agree excellently with the evolution of the Allen-Cahn equation.

\bibliographystyle{plain}
\bibliography{references}

\end{document}